\documentclass[twocolumn, switch]{article} % Method A for two-column formatting

\usepackage{preprint}

\usepackage{amsmath, amsthm, amssymb, amsfonts}

\usepackage[numbers,square]{natbib}
\usepackage[utf8]{inputenc}	% allow utf-8 input
\usepackage[T1]{fontenc}	% use 8-bit T1 fonts
\usepackage{xcolor}		% colors for hyperlinks
\usepackage[colorlinks = true,
            linkcolor = purple,
            urlcolor  = blue,
            citecolor = cyan,
            anchorcolor = black]{hyperref}	% Color links to references, figures, etc.
\usepackage{booktabs} 		% professional-quality tables
\usepackage{nicefrac}		% compact symbols for 1/2, etc.
\usepackage{microtype}		% microtypography
\usepackage{lineno}		% Line numbers
\usepackage{float}			% Allows for figures within multicol
\usepackage{makecell}
\usepackage[normalem]{ulem}
\usepackage{multirow}
\usepackage{xurl}

\usepackage{newfloat}
\DeclareFloatingEnvironment[name={Supplementary Figure}]{suppfigure}
\usepackage{sidecap}
\sidecaptionvpos{figure}{c}

\usepackage{titlesec}
\titlespacing\section{0pt}{12pt plus 3pt minus 3pt}{1pt plus 1pt minus 1pt}
\titlespacing\subsection{0pt}{10pt plus 3pt minus 3pt}{1pt plus 1pt minus 1pt}
\titlespacing\subsubsection{0pt}{8pt plus 3pt minus 3pt}{1pt plus 1pt minus 1pt}

\title{Measuring the environmental impact of delivering AI at Google Scale}

\usepackage{authblk}

\author{Cooper Elsworth}
\author{Keguo Huang}
\author{David Patterson}
\author{Ian Schneider}
\author{Robert Sedivy}
\author{Savannah Goodman}
\author{Ben Townsend}
\author{Parthasarathy Ranganathan}
\author{Jeff Dean}
\author{Amin Vahdat}
\author{Ben Gomes}
\author{James Manyika}

\affil{Google, Mountain View, CA, USA}

\begin{document}

\twocolumn[ % Method A for two-column formatting
  \begin{@twocolumnfalse} % Method A for two-column formatting
  
\maketitle

\begin{abstract}
The transformative power of AI is undeniable---but as user adoption accelerates, so does the need to understand and mitigate the environmental impact of AI serving. However, no studies have measured AI serving environmental metrics in a production environment. This paper addresses this gap by proposing and executing a comprehensive methodology for measuring the energy usage, carbon emissions, and water consumption of AI inference workloads in a large-scale, AI production environment. Our approach accounts for the full stack of AI serving infrastructure---including active AI accelerator power, host system energy, idle machine capacity, and data center energy overhead. Through detailed instrumentation of Google's AI infrastructure for serving the Gemini AI assistant, we find the median Gemini Apps text prompt consumes 0.24 Wh of energy---a figure substantially lower than many public estimates. We also show that Google’s software efficiency efforts and clean energy procurement have \textcolor{black}{driven a 33x reduction in energy consumption and a 44x reduction in carbon footprint for the median Gemini Apps text prompt over one year.} We identify that the median Gemini Apps text prompt uses less energy than watching nine seconds of television (0.24 Wh) and consumes the equivalent of five drops of water (0.26 mL). While these impacts are low compared to other daily activities, reducing the environmental impact of AI serving continues to warrant important attention. Towards this objective, we propose that a comprehensive measurement of AI serving environmental metrics is critical for accurately comparing models, and to properly incentivize efficiency gains across the full AI serving stack.
\end{abstract}
%\keywords{First keyword \and Second keyword \and More} % (optional)
\vspace{0.35cm}

  \end{@twocolumnfalse} % Method A for two-column formatting
] % Method A for two-column formatting

%\begin{multicols}{2} % Method B for two-column formatting (doesn't play well with line numbers), comment out if using method A

%%%%%%%%%%%%%%%  Main text   %%%%%%%%%%%%%%%
\section{Introduction} \label{sec:introduction}
\textit{Artificial intelligence} (AI) is reshaping industries and daily life, driven largely by the accelerating capabilities of \textit{Large Language Models} (LLMs). While much of the initial focus on AI’s environmental impact rightly centered on the energy-intensive process of model training~\citep{strubell2020energy,patterson2021carbon,luccioni2023counting}, the surge in public adoption of generative AI applications has shifted attention toward the footprint of AI model inference and serving. With these AI models now serving billions of user prompts globally, the energy, carbon emissions, and water impacts associated with generating responses at scale represents a significant and rapidly growing component of AI's overall environmental cost~\citep{gupta2021chasing,patterson2022carbon}.

In response, several research efforts and disclosures have emerged to quantify the per-prompt energy consumption of inference (Wh/prompt). Early work provided coarse estimates of energy consumption per prompt, relying on high-level assumptions about hardware specifications and model parameters~\citep{de2023growing}. More recently, initiatives like the the AI Energy Score~\citep{luccioni2025ai} and the ML.ENERGY~\citep{chung2025ml} benchmarks have advanced the field by employing direct empirical measurements. These frameworks aim to standardize energy transparency by benchmarking models on specific tasks using consistent hardware. In addition, other studies have expanded the aperture to consider the carbon emissions and water consumption associated with serving AI models. 

Despite this progress, the field lacks first-party data from the largest AI model providers. Based on decades of deploying software at scale, \textcolor{black}{Google has a unique perspective on the operational realities of maintaining a large-scale, globally-distributed AI production fleet, and serving software products at scale---such as web search. Characterizing and optimizing the environmental impact of AI model serving requires a comprehensive view} of energy consumption---including the power drawn by the host machine’s CPU and DRAM, the significant energy consumed by idle systems provisioned for reliability and low latency, and the full data center overhead as captured by the \emph{Power Usage Effectiveness} (PUE) metric~\citep{avelar2012pue}. \textcolor{black}{The missing consensus on the energy-consuming activities to include in the measurement---known as the \textit{measurement boundary}---has} led to published estimates for similar AI tasks varying by an order of magnitude. A lack of agreed upon methodologies may have contributed to a lack of first-party data when it is needed most~\citep{luccioni2025misinformation}.

\textbf{Contributions} This paper \textcolor{black}{presents the energy, emissions, and water impacts for a production AI product} by establishing a comprehensive framework to measure critical aspects of serving AI at Google's scale. First, we propose a full-stack measurement approach that accounts for all material energy sources. Second, we apply this methodology to Google's Gemini Apps product to provide the first analysis of three AI serving environmental metrics:
\begin{itemize}
    \item \textit{Energy / prompt:} the energy consumption required to serve an AI assistant text prompt
    \item \textit{Emissions / prompt:} the \textit{market-based} (MB) emissions generated by grid electricity generation (including renewable energy procurement), and the embodied emissions of the AI accelerator hardware 
    \item \textit{Water consumption / prompt:} the water consumed for cooling machines and associated infrastructure in data centers
\end{itemize}
\textcolor{black}{We demonstrate that existing---and often narrower---measurement approaches are missing material energy consumption activities for AI serving.} Finally, we illustrate the compounding AI serving efficiency gains across the serving stack over a year of development, resulting in a 44x reduction in the total emissions generated for the median Gemini Apps prompt. Comprehensive environmental metrics---like those proposed in this paper---are critical to properly incentivize efficiency opportunities across a large-scale, globally distributed production fleet.

\section{Related Work} \label{sec:related-work}
Efforts to quantify the environmental impact of AI inference can be broadly categorized into two approaches: \textit{model-based estimation} and \textit{empirical measurement}. Early work primarily rely on estimation, calculating energy consumption from publicly available hardware specifications, known model parameters, and a series of assumptions regarding usage patterns like prompt complexity and token length~\citep{IEA2025EnergyAI}. While these theoretical models are helpful for highlighting the potential scale of the problem, their results are highly sensitive to the underlying assumptions and may suffer from accumulating errors in the estimated values. 

In contrast, more recent initiatives have shifted toward empirical measurement approaches, using software tools~\citep{zeus-nsdi23,benoit_courty_2024_11171501} to directly measure energy consumption on standardized hardware during the execution of specific tasks. Moving to more empirical metrics has increased the precision of these metrics, but still suffers from large differences in the underlying methodologies and comparability between studies. In this section, we outline existing approaches to make the case for a more consistent and complete measurement approach.

\begin{figure*} [t]
	\centering
	\includegraphics[width=.8\textwidth]{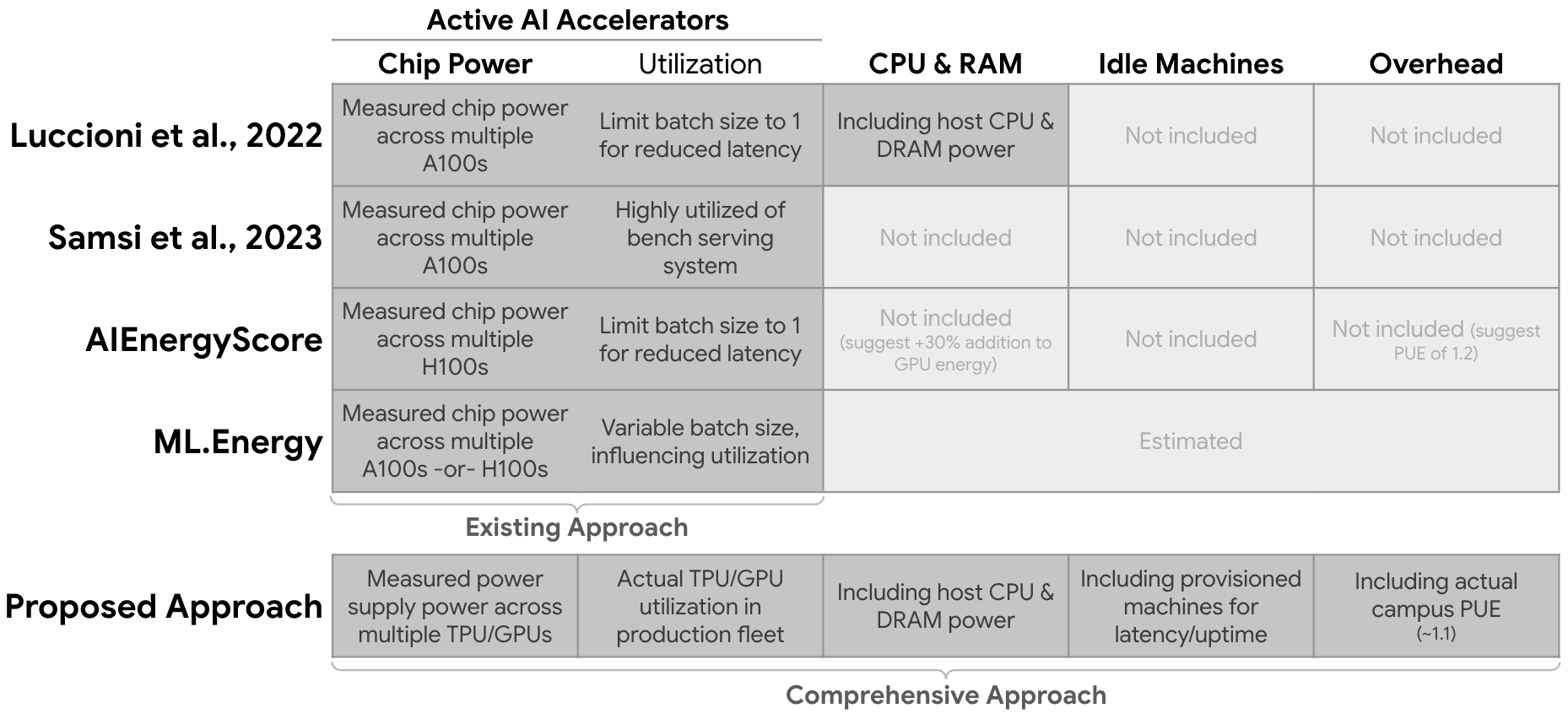}
	\caption{Existing and proposed boundaries for AI inference energy measurements. The existing approaches primarily consider the energy consumption of the active AI Accelerators. We propose including measured energy from all components of the serving stack to provide a more comprehensive measurement of AI inference energy consumption to inform reduction levers across the serving stack.}
	\label{fig:related}
\end{figure*}

\subsection{Estimated metrics}
Model-based estimations of AI inference environmental metrics are widespread and variable, due to their sensitivity on poorly-constrained input assumptions. Some of the most referenced and recent studies include:
\begin{itemize}
    \item \textit{De Vries, 2023~\citep{de2023growing}}: This work takes a model-based estimation approach to calculate prompt energy use based on publicly available hardware specifications (e.g., NVIDIA A100 GPUs), AI model parameters (e.g., 175B for GPT-3.5), and critical assumptions about typical usage patterns, such as input/output token lengths for a prompt. These results suggest that a single GPT-3.5 prompt may consume around 3 \textit{watt-hours} (Wh) of energy.
    \item \textit{Epoch.AI, 2025~\citep{epoch2025howmuchenergydoeschatgptuse}}: This analysis employs a model-based estimation methodology and updates several key assumptions to reflect more modern AI hardware and usage patterns. The calculation assumes that the prompt was processed by the GPT-4o model, which is believed to use a mixture-of-experts architecture with about 100 billion active parameters, running on NVIDIA H100 GPUs. The energy consumption of a typical ChatGPT prompt is estimated to be approximately 0.3 Wh.
    \item \textit{EcoLogits~\citep{rince2025ecologits}}: This study defines a regression model for GPU power based on LLM Performance, CPU and DRAM model, and applies average data center PUE. For a small (50 output token) prompt, the EcoLogits calculator estimates a range from 1.83 Wh to 6.95 Wh.
    \item \textit{Li et al., 2025~\citep{li2025makingaithirstyuncovering}}: Researchers estimate that inference for the GPT-3 model in Microsoft's U.S. data centers consume a 500 mL bottle of water for roughly 10 to 50 medium-length responses, which translates to approximately 10-50 mL per prompt.
    \item \textit{Sam Altman, 2025~\citep{altman2025}}: In a June 2025 blog post, OpenAI CEO Sam Altman disclosed that an average ChatGPT prompt consumes approximately 0.34 Wh of energy and a very small amount of water (0.000085 gallons, or 0.3 mL). The disclosure provides no explanation of the measurement boundary or methodology used to arrive at this number, making it impossible to compare with other estimates or to understand which components of the serving stack were included.
    \item \textit{Mistral AI, 2025~\citep{mistral2025}}: A peer-reviewed \textit{lifecycle assessment} (LCA) for its Mistral Large 2 model was conducted in collaboration with the French environmental agency ADEME and consulting firm Carbone 4. For a typical 400-token response from its "Le Chat" assistant, Mistral reports a marginal impact of  1.14 grams of CO$_2$e, and 45 milliliters (mL) of water consumed.
\end{itemize}
\textcolor{black}{There is an order-of-magnitude of variability in the estimated energy per chat prompt, which makes it difficult for a user to understand the environmental impact of using an AI assistant. Moreover, a reliance on estimated environmental impacts limits our ability to identify and implement environmental impact reductions.}

\subsection{Measured metrics}
\begin{figure*}[t]
	\centering
	\includegraphics[width=.75\textwidth]{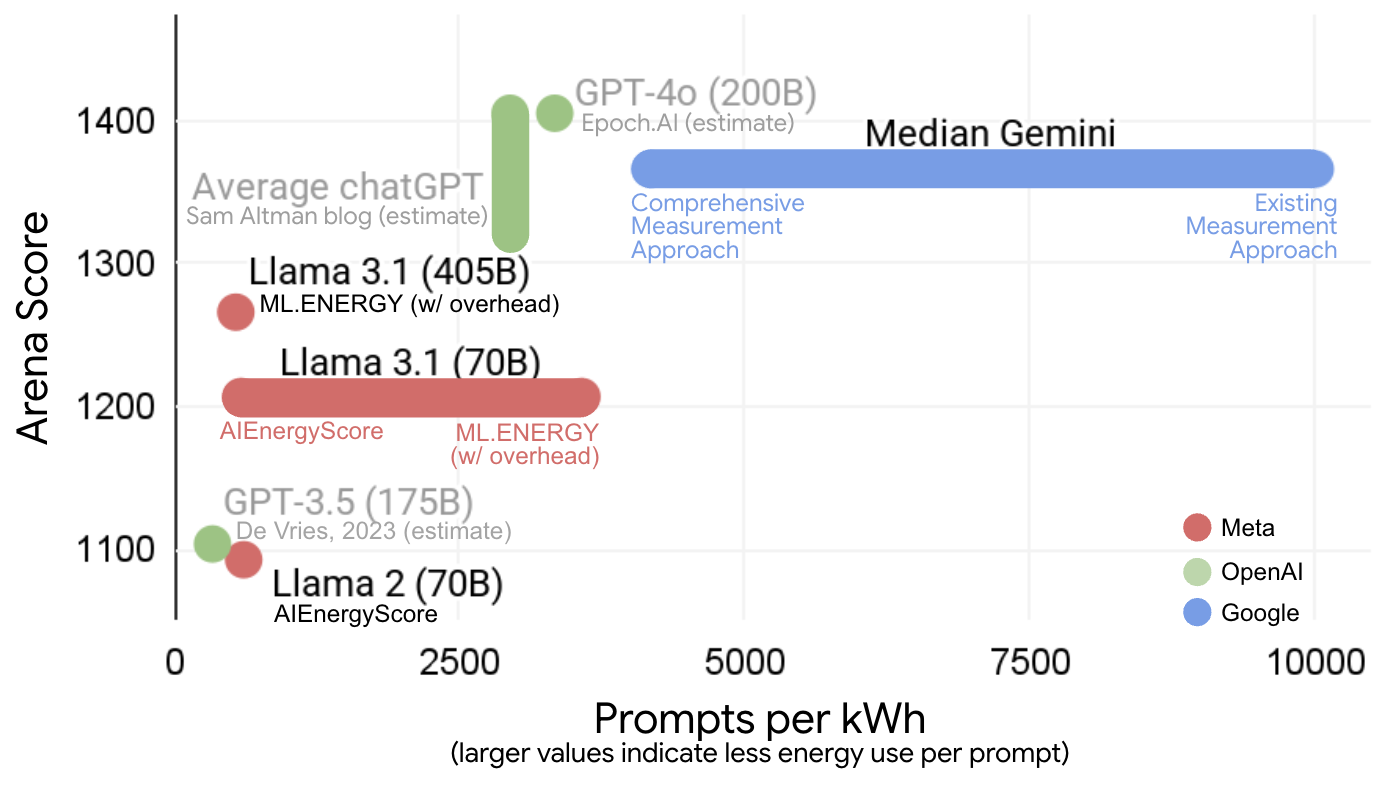}
	\caption{Energy per prompt results for large production AI models plotted against LMArena score~\citep{chiang2024chatbot} -- estimated metrics (gray text) and measured metrics (black, red, and blue text). For similar models and parameter counts, the different measurement approaches can lead to large differences in results, e.g. Llama 3.1 (70B) varies from \textasciitilde 580 to \textasciitilde 3600 prompts per kWh. The results for the median Gemini Apps text prompt presented in this paper vary from a more narrowly defined Existing Approach (10,000 prompts per kWh) to the more complete Comprehensive Approach (4,167 prompts per kWh) proposed.}
	\label{fig:other-models}
\end{figure*}

There have been four primary studies directly measuring energy/prompt metrics from AI accelerator hardware. However, these studies exhibit notable differences in how they measure energy consumption, which limits comparability. Figure~\ref{fig:related} illustrates the differences in methodology, and in summary:
\begin{itemize}
    \item \textit{Luccioni et al., 2022~\citep{luccioni2023estimating}}: This study provides estimates for the BLOOM model's energy and emissions footprint during inference. Using the CodeCarbon library~\citep{benoit_courty_2024_11171501}, this measurement includes GPU, CPU and DRAM energy consumption. The average metrics over 18 days of inference with no batched inference are 4 Wh/prompt and 1.5 gCO$_2$e/prompt.
    \item \textit{Samsi et al., 2023~\citep{samsi2023words}}: This research benchmarks several sizes of the LLaMA model (7B, 13B, and 65B parameters) on two generations of NVIDIA GPUs (V100 and A100). For the LLaMA-65B model, they measure an energy consumption of approximately 0.3 Wh per response.
    \item \textit{AI Energy Score~\citep{luccioni2025ai}}: This initiative employs empirical benchmarking of models, and conducts direct energy measurements on standardized hardware (specifically NVIDIA H100 GPUs). The primary tool for energy measurement is CodeCarbon~\citep{benoit_courty_2024_11171501} with a focus on GPU energy consumption to determine a comparative star rating. AI Energy Score aims to standardize energy transparency by benchmarking models on specific tasks using consistent hardware. Prioritizing comparability, they hold the inference batch size to 1, which is likely not representative of a production inference environment. We expect the batch size constraint to reduce AI accelerator utilization, increasing the energy/prompt metrics.
    \item \textit{ML.ENERGY~\citep{chung2025ml}}: This framework also relies on empirical benchmarking using production-grade hardware like NVIDIA H100s. It places a strong emphasis on emulating real-world serving conditions, including steady-state operation and sophisticated handling of batching. GPU energy is measured using the Zeus library~\citep{zeus-nsdi23}. The goal of this benchmark is to determine per-request energy figures that accurately reflects operational deployment scenarios.  ML.ENERGY Benchmark emphasizes more realistic inference conditions in production, measuring per-request energy by considering factors like steady-state operation and batching, aiming for actionable optimization insights. Coverage of ML.ENERGY results in the MIT Technology Review~\citep{ai_energy_2025} applied a doubling of energy per prompt to estimate overhead energy (cooling, other computations, and other demands). 
\end{itemize}
The move from model estimation to empirical measurement represents a significant step toward greater accuracy and transparency. However, the specific boundaries of these measurements are critical for interpreting their results. For example, two of the aforementioned measurement approaches, with distinct assumptions and boundaries, found a 6x difference in the energy/query of Llama 3.1 (with 70B parameters) (Figure~\ref{fig:other-models}).

As we explore in this study, a large-scale inference system can vary significantly from the benchmarks seen in the literature. Most notably, the utilization of AI accelerators directly impacts the energy consumption per unit of computation, and large-scale inference systems strive to maximize compute efficiency. Examples of approaches to improve AI serving compute efficiency (many of which Google has pioneered or implemented at scale) include:
\begin{enumerate}
    \item \textit{Batch Inference} allows for multiple inference prompts to be handled concurrently to maximize the utilization of the AI accelerators. Smaller batch size prioritizes latency, while a larger batch size prioritizes throughput and efficiency~\citep{chung2025ml}.
    \item \textit{Speculative Decoding} employs a small embedded draft model for decoding, and verifies the result with the larger model. Successful speculation skips a significant amount of the full model decoding steps, leading to more efficient inference~\citep{leviathan2023fast}.
    \item \textit{Disaggregated Serving} places Transformer's prefill and decoding on separate accelerators, and optimizes each computation separately. This split allows for more efficient prefill and decoding steps with lower latency and higher throughput~\citep{hu2024memserve}. 
    \item \textit{KV Caching} is an optimization technique that makes Transformer models more efficient by saving interim results. It works by saving the calculated key and value matrices from the attention mechanism for previous tokens, which avoids redundant computations when generating the next token~\citep{li2024survey}.
    \item \textit{Optimized Software-Hardware Stack} allows for optimizations across stack layers of LLM serving, including compilation/kernel optimizations that make things more efficient than what external studies might assume for their models. For example, Google's XLA ML compiler~\citep{openxla_xla}, Pallas kernels~\citep{jaxpallas}, and Pathways systems~\citep{barham2022pathways} enable model computations expressed in higher-level systems like JAX to be run efficiently on accelerator serving hardware.
\end{enumerate}

\textcolor{black}{
Relative to earlier work, this study provides a production fleet perspective on the realities of serving AI products. In the following sections, we strive to define AI serving environmental metrics that:
\begin{enumerate}
    \item allow meaningful comparison between scaled AI Products
    \item Set a reasonably broad standard for the measurement, boundary, to encourage future work to accurately and comprehensively measure environmental costs, and
    \item Incentivize actions to optimize the full-stack of energy-consuming activities.
\end{enumerate}
However, we understand that this contribution does not address all priorities of previous studies---which include controlling for hardware differences, ease of measurement, or relative visibility into layers of the serving stack.
}

\section{Methodology} \label{sec:methodology}
This paper proposes a comprehensive approach to measure energy usage, emissions generation, and water consumption for serving production AI models at scale. We believe that this comprehensive approach can form a consistent standard that accurately measures AI serving carbon emissions and aligns incentives for emissions reduction.  The methodology has four parts: 1) a comprehensive measurement boundary and transparent exclusions, 2) energy measurement methodology, 3) emissions and water measurement methodologies, and 4) choice of an aggregate representative metric. 

\subsection{Measurement Boundary \& Exclusions}
\textcolor{black}{The measurement boundary is a definition of the activities that are included in metrics. In this case, the measurement boundary includes energy consuming activities associated with LLM serving, which drives emissions generation and water consumption.} We consider the measurement boundary for LLM serving energy consumption to include material energy sources under Google's operational control---i.e. the ability to implement changes to behavior. The functional unit for this study is one serving AI computer deployed in the data center, which includes one or more AI accelerator trays (containing AI accelerators) connected to one host tray. Specifically, we decompose energy consumption as:
\begin{enumerate}
    \item \textit{Active AI Accelerator energy:} This metric includes the energy consumed by all AI accelerators connected to the active AI computer, including Tranformer prefill and decode. This energy consumption includes the networking communication between AI accelerators in the same AI computer. This result is based on direct measurements during serving, so it accounts for actual accelerator utilization in a production system.
    \item \textit{Active CPU \& DRAM energy:} A host tray’s CPU and DRAM are necessary to run the accelerators, and both of these components are considered within scope of the energy consumption. This metric includes energy consumed by the CPU and DRAM on the active AI computer.
    \item \textit{Idle Machine energy:} To ensure high availability and low latency for users globally, production systems require a reserved capacity that may be idle at any given moment but is ready to handle traffic spikes or failover. In addition, systems may have temporary idle states during workload transitions. The energy consumed by these idle AI computers, which are essential for service uptime, must be factored into the total energy footprint. This metric includes the energy consumed by idle AI accelerators and associated host trays that are provisioned to serve the relevant model and product.
    \item \textit{Overhead energy:} The infrastructure supporting data centers---including cooling systems, power conversion, and other overhead within the data center---also consumes energy. This overhead is captured by the \textit{Power Usage Effectiveness} (PUE) metric~\citep{avelar2012pue}. This metric includes the energy consumption associated with overhead data center activities. 
\end{enumerate}

\textcolor{black}{The definition of the serving AI computer functional unit leads to a few} notable exclusions to the measurement boundary that include: 
\textcolor{black}{
\begin{itemize}
    \item \textit{Networking:} AI model prompts arrive at the edge of Google's network through external networking, and are routed to and dispatched to an AI computer through data center networking. External networking energy consumption is excluded due to a lack of operational control, and data center networking energy is estimated to be negligible for an AI assistant text prompt.
    \item \textit{End user devices:} Energy consumption of end user devices, including edge compute, is excluded due to a lack of operational control on these devices. 
    \item \textit{LLM training \& data storage:} This study specifically considers the inference and serving energy consumption of an AI prompt. We leave the measurement of AI model training to future work.
\end{itemize}
}

\subsection{Energy Measurement Methodology}\label{sec:energy-method}
\begin{figure}[t]
	\centering
	\includegraphics[width=.45\textwidth]{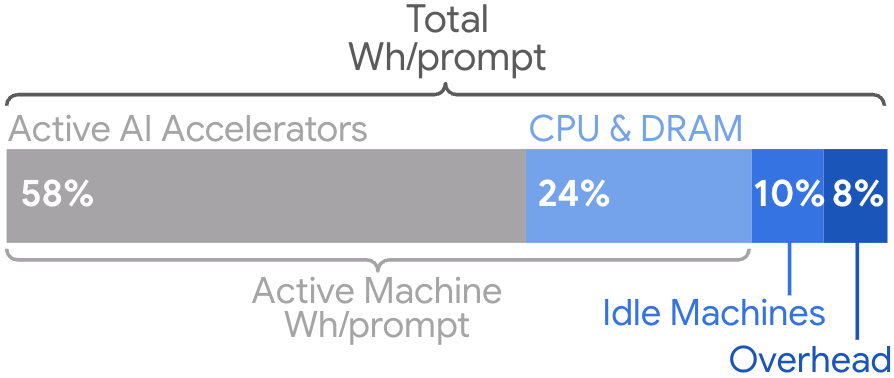}
	\caption{Components of the total LLM energy consumption per prompt across a production LLM serving stack. The relative size of the energy components is based on the median Gemini Apps text prompt in May 2025 using the Comprehensive Measurement approach.}
	\label{fig:methodology-components}
\end{figure}

To measure the in-situ fleet environmental metrics of a Gemini Apps prompt, we have developed a methodology following the measurement boundary outlined in Section 3.1. We denote this proposed methodology as the Comprehensive Approach, and it is \textcolor{black}{based on internal telemetry deployed across Google's AI serving fleet. The approach follows:}
\begin{enumerate}
    \item Identify all LLM models serving the Gemini app, including all supporting models for scoring, ranking, classification, and other prompt routing tasks.
    \item Map LLM models to the \textcolor{black}{job IDs (i.e. a set of tagged computations) and machine IDs (i.e. AI computers  assigned to serve those jobs)} to measure \textit{power supply unit} (PSU) energy consumption for the machines during inference runtime, following the underlying energy measurement approach from~\citet{schneider2024carbon}.
    
    \textcolor{black}{Specifically, we collect tray power, $P$, measurements for the host and AI accelerators based on external \textit{power supply units} (PSUs). Concurrently, we track the total time each machine is allocated to a given job ($t_{total}$) and the portion of that time it spends idle ($t_{idle}$), where $t_{idle} <t_{total}$.}
    Therefore, the energy components for a given model shown in Figure~\ref{fig:methodology-components} are defined as:
    \begin{align*} 
        E_{Total} &= \sum_\text{machine, hour} P_{total} * t_{total} * PUE \\ 
        E_{Overhead} &= \sum_\text{machine, hour} P_{total} * t_{total} * (PUE - 1) \\
        E_{Idle} &= \sum_\text{machine, hour} P_{idle} * t_{idle} \\
        E_{Active Machines} &= E_{Total} - E_{Overhead} - E_{Idle} \\
        E_{Active CPU \& DRAM} &= E_{Active Machines} * P_{host} / P_{total} \\
        E_{Active AI Accelerators} &= E_{Active Machines} * P_{accel} / P_{total}
    \end{align*}
    where 
    \begin{itemize}
        \item $E_{x}$: An individual energy component $x$ ($x$ = total, overhead, idle, active CPU \& DRAM, active AI Accelerators)
        \item $t_{total}$: Total time allocated to the model.
        \item $t_{idle}$: The time that the machine is idle. The active time is therefore $t_{total}$ - $t_{idle}$.
        \item $P_{idle}$: The machine's baseline power consumption when idle. 
        \item $P_{host}$: Hourly average power consumed by the host system (CPU \& DRAM).
        \item $P_{accel}$: Hourly average power consumed by the AI Accelerators.         \item $P_{total}$: Hourly average total machine power, given by the sum
$P_{total} = P_{host} + P_{accel}$.
        \item $PUE$: Campus-level power utilization effectiveness.
    \end{itemize}
    
    \item Measure total user prompt count across Gemini Apps products for each LLM model, denoted as $Q$.
    \item Divide each energy component, $E_{x}$, by total user prompt count over the same measured time period to get energy/prompt,
    \begin{align*}
        E_{x/prompt} = \frac{E_{x}}{Q}
    \end{align*}
    We measure over a day or longer. This is sufficiently longer than the prompt duration, so partial prompt counts do not measurably impact the results.
\end{enumerate}

\begin{table*}[t] 
 \caption{Energy consumption of the median Gemini Apps text prompt (Wh/prompt) in May 2025 using the Existing and Comprehensive methodologies illustrated by Figure~\ref{fig:related}. In the more narrow Existing Approach, CPU \& DRAM, Idle Machines, and Overhead energy consumption are not included---so they are presented, but not included in the total energy.}
  \centering
  \begin{tabular}{rccccccc}
    & \multicolumn{2}{c}{Active AI Accelerators} & & & & \multirow{3}{*}{\textbf{\makecell{Gemini \\ Wh/prompt}}}\\
    \cmidrule(r){2-3}
    & AI Accel. Power & Util. sample & CPU \& DRAM & Idle Machines & Overhead &  \\
    \cmidrule(r){2-6}
    \makecell[r]{Existing \\ Approach} & 0.10  & \makecell{Prompts in Top 10\% \\ most efficient DCs} & \textcolor{gray}{\sout{0.04}} & \textcolor{gray}{\sout{0.02}} & \textcolor{gray}{\sout{0.01}} & \textbf{0.10} \\
    \cmidrule(r){2-6}
    \makecell[r]{Comprehensive \\ Approach} & 0.14 & Average across fleet & 0.06 & 0.02 & 0.02 & \textbf{0.24} \\
    \cmidrule(r){2-6}
  \end{tabular}
  \label{tab:energy-metrics}
\end{table*}

\textcolor{black}{In contrast to the proposed Comprehensive Approach, we also define an Existing Approach that more narrowly considers a highly utilized benchmark being run on a standalone AI accelerator. This Existing Approach is more comparable to that found in the existing literature. It differs from the Comprehensive Measurement in two ways. First, it measures $E_{Active AI Accelerators}$ instead of $E_{Total}$. Second, it subsamples energy consumption data from the top 10\% most energy-efficient data centers, defined by lowest daily average energy/prompt. The higher effective utilization from this subsampling mimics a highly utilized benchmark study. This and the more narrow boundary allow for a more direct comparison with existing benchmarks in the literature (see Figure~\ref{fig:related}).}

\subsection{Emissions \& Water Metrics Methodology}\label{sec:water-emission}
\textit{Emissions:} Carbon emissions are generated based on the local grid energy mix of the consumed electricity, and the embodied emissions of the compute hardware. Therefore, we calculate the carbon emissions of LLM serving as,
\begin{align*}
    CO_2e_{/prompt} = E_{Total/prompt} * EF + (Scope1 + Scope3)/Q
\end{align*}
where $EF$ is the previous calendar-year's average annual grid emission factors across Google data centers, following \textit{market-based} (MB) standards from the greenhouse-gas (GHG) protocol~\citep{sotos2015ghg}. \textit{Market-based} (MB) GHG emissions account for the GHG emissions emitted by generating sources from which a company purchases electricity and associated environmental attributes. Therefore MB emissions credit companies for carbon-free energy purchases, allowing them to reduce their associated MB emissions. Alternatively, \textit{location-based} (LB) GHG emissions refer to the GHG emissions emitted within a specific geographic boundary, such as a country or grid region, and exclude the impact of a company's CFE procurement. We consider MB GHG emissions for all metrics presented in this paper.

For 2023~\citep{google2024environmental}, Google’s LB emission factor was 366 gCO2e/kWh and the reduction associated with its procured CFE was equivalent to 231 gCO2e/kWh, so the net MB emission factor was 135 gCO2e/kWh. For 2024~\citep{google2025environmental}, Google’s LB emission factor was 345 gCO2e/kWh and the reduction associated with its procured CFE was equivalent to 251 gCO2e/kWh, so the net MB emission factor was 94 gCO2e/kWh. We apply the previous calendar-year's average emissions factors to account for the fact that these metrics are published once per year.

For metrics on carbon emissions,~\citet{schneider2025life} highlight the need to include full lifecycle emissions---including operational (Scope 1 \& 2) and embodied (Scope 3) contributions. While including embodied (Scope 3) emissions is not common in the literature~\citep{schneider2025life}, we include it in this study to be as comprehensive as possible in the measurement boundary. $Scope1$, $Scope3$ are the associated Scope 1 and Scope 3 GHG emissions for the AI accelerators and host CPU \& DRAM based on results in~\citet{schneider2025life}.
    
\textit{Water Consumption:} Google’s data centers often rely on water for cooling to reduce overhead energy consumption. Water consumption efficiency is measured using the \textit{Water Usage Effectiveness} (WUE) metric, specifically the consumptive use variant (ISO WUE Category 2) calculated as water input minus water returned. Though we calculate both the withdrawal and consumption intensity, we consider consumption to best represent our impact on local water availability, as it represents the volume of water that is evaporated and therefore unavailable for reuse. On average, Google consumes 80\% of the water withdrawn. We calculate the water consumption of LLM serving as a function of energy/prompt as,
\begin{align*}
    Water_{/prompt} = (E_{Total/prompt} - E_{Overhead/prompt}) * WUE
\end{align*}
where $WUE$ is the previous calendar-year average freshwater \textit{Water Usage Effectiveness} Category 2~\citep{azevedo2011water} of Google's data centers supporting LLM models to normalize for seasonal or site-specific variation. For both 2023 and 2024, Google’s WUE Category 2 value was 1.15 L/kWh.

\subsection{Aggregate Metric Definition}
We attempt to define aggregate energy, emissions, and water metrics that are representative of a typical user's behavior and comparable over time. However, we recognize that this can be difficult with the rapidly evolving landscape of AI model architectures and AI assistant user behavior. We find that the distribution of energy/prompt metrics can be skewed, with the skewed outliers varying significantly over time. Part of this skew is driven by small subsets of prompts served by models with low utilization or with high token counts, which consume a disproportionate amount of energy. In such skewed distributions, the arithmetic mean is highly sensitive to these extreme values, making it an unrepresentative measure of typical user's impact. In contrast, the median is robust to extreme values and provides a more accurate reflection of a typical prompt's energy impact.  Consequently, we use the daily median as the aggregate value for the metrics defined in Section~\ref{sec:energy-method} and Section~\ref{sec:water-emission}. 

To calculate the energy consumption for the median Gemini Apps text prompt on a given day, we first determine the average energy/prompt for each model, and then rank these models by their energy/prompt values. We then construct a cumulative distribution of text prompts along this energy-ranked list to identify the model that serves the 50-th percentile prompt. The average energy/prompt is defined as the energy of the median Gemini Apps text prompt on that day, and the monthly median energy/prompt is calculated as described above for the corresponding month. Finally, the corresponding carbon emissions and water consumption are derived by applying the conversion factors defined in Section~\ref{sec:water-emission} on these median energy metrics.

\section{Results}
This section presents the environmental impact metrics for the Gemini Apps AI assistant, calculated using both an existing, narrower measurement standard and our proposed comprehensive methodology (see Figure~\ref{fig:related}). The results highlight the significant differences between the two approaches and underscore the importance of a full-stack measurement framework.

\subsection{Gemini Energy Consumption}
Applying our comprehensive measurement methodology, we find that the median energy consumption for a Gemini Apps text prompt in May 2025 is 0.24 Wh. As Table~\ref{tab:energy-metrics} and Figure~\ref{fig:methodology-components} show, the primary energy draw originates from the active AI Accelerator power (0.14 Wh, 58\% of total) and the necessary host CPU \& DRAM power (0.06 Wh, 25\%). The energy consumed by provisioned idle machines and the data center overhead (PUE) each contribute 0.02 Wh (10\% and 8\% respectively). \textcolor{black}{This suggests that a scaling of 1.72 would need to be applied to active AI accelerator energy consumption to include the energy consumed in a production serving environment, compared to a 2 times scaling from existing estimates~\citep{ai_energy_2025}.}

In contrast, when applying a methodology aligned with \textcolor{black}{a more narrow, existing approach} (Figure~\ref{fig:related}), the calculated energy consumption is only 0.10 Wh per prompt. This lower figure results from both the more limited measurement boundary, and considering a sample of more highly utilized machines (similar to an idealized benchmark study where utilization is highly optimized). The comprehensive approach reveals a total energy consumption that is 2.4 times greater than the estimate from the existing approach, demonstrating the need to standardize measurement approaches and boundaries to be more inclusive of all energy consumptive activities.

Both approaches show a per-prompt energy consumption figure that is lower than many results presented in the literature (see Figure~\ref{fig:other-models}). See Section 4.3 for a discussion of likely contributors.

\subsection{Gemini Emissions \& Water Consumption}
While energy per-prompt is an important factor to consider and the most directly comparable between models, environmental impact ultimately comes from generating emissions and consuming water. Google's efforts to reduce the emissions intensity (gCO$_2$e/kWh) and limit water use in high-stress watersheds have significantly reduced the impact of AI serving on the environment~\citep{google2023water}.

\begin{table}[h] 
 \caption{Energy, emissions, and water usage of the median Gemini Apps text prompt in May 2025 using the existing and proposed approaches described in Section 3.2.}
  \centering
  \begin{tabular}{rcc}
    & \makecell{Existing \\ Approach} & \makecell{Comprehensive \\ Approach} \\
    \cmidrule(r){2-3}
    Energy (Wh/prompt) & 0.10 & 0.24  \\
    \cmidrule(r){2-3}
    Emissions (gCO$_2$e/prompt) & 0.02 & 0.03 \\
    (Scope 2 MB) & \multicolumn{1}{r}{(0.016)} & \multicolumn{1}{r}{(0.023)} \\
    (Scope 1+3) & \multicolumn{1}{r}{(0.007)} & \multicolumn{1}{r}{(0.010)} \\
    \cmidrule(r){2-3}
    Water (mL/prompt) & 0.12 & 0.26 \\
    \cmidrule(r){2-3}
  \end{tabular}
  \label{tab:co2-water-metrics}
\end{table}

As Table~\ref{tab:co2-water-metrics} shows, 
a median Gemini Apps text prompt generates 0.03 gCO$_2$e and consumes 0.26 mL of water when measured comprehensively. These values are even lower when considering a methodology more similar to a narrow, existing approach (0.01 gCO$_2$e and 0.15 mL of water consumption). To put this into context: a modern television consumes approximately 100 watts of electricity~\citep{Marsh2024}, so 0.24 Wh represents less energy than watching TV for 9 seconds. The water use of 0.26 mL equals five drops of water (based on a standard 0.05 mL drop), orders of magnitude less than previous estimates of 45~\citep{mistral2025} to 50 mL~\citep{li2025makingaithirstyuncovering}.

\begin{figure}[t]
	\centering
	\includegraphics[width=.5\textwidth]{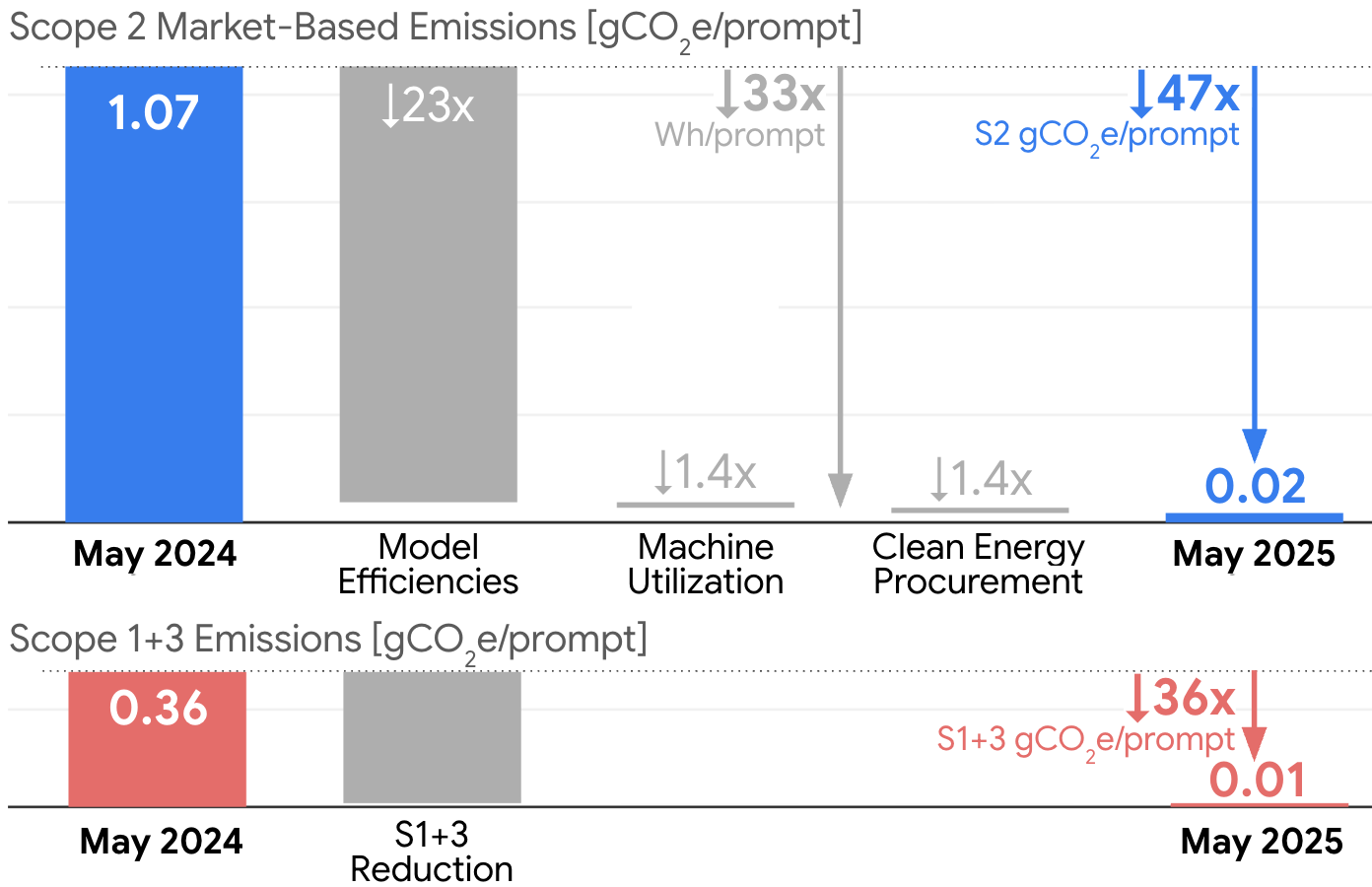}
	\caption{Median Gemini Apps text prompt emissions over time---broken down by Scope 2 MB emissions (top) and Scope 1+3 emissions (bottom). Over 12 months, we see that AI model efficiency efforts have led to a 47x reduction in the Scope 2 MB emissions per prompt, and 36x reduction in the Scope 1+3 emissions per user prompt---equivalent to a 44x reduction in total emissions per prompt.}
	\label{fig:efficiency}
\end{figure}

Google’s Water Risk Framework adopted in 2023, and our commitment to Responsible Water Use~\citep{google2023water}, ensures that all new data centers developed in high-stress locations will make use of air-cooled technology during normal operations. As a result, we anticipate our water usage effectiveness in high-stress areas will trend significantly lower than the fleet average and approach zero as legacy assets reach end of life. Improvements in AI serving energy efficiency and water usage effectiveness will compound to further reduce the water consumption impact of AI assistant prompts.  

The breakdown of the total per-prompt emission between market based electricity-related emissions and embodied emissions remains consistent with results presented in~\citet{schneider2025life}, where electricity-related emissions dominate. This informs that the most impactful reduction initiatives are energy efficiency improvements and decarbonization of electricity consumed.

\subsection{Gemini Emissions Efficiency Gains}
A key motivation for the comprehensive measurement of AI serving environmental metrics is to track and incentivize optimizations across the entire serving stack. By tracking these metrics from May 2024 to May 2025, we demonstrate a 44x reduction in the total emissions per median Gemini Apps text prompt over 12 months (Figure~\ref{fig:efficiency}). This impact results from:
\textcolor{black}{
\begin{enumerate}
    \item A 33x reduction in per-prompt energy consumption driven by software efficiencies---including a 23x reduction from model improvements, and a 1.4x reduction from improved machine utilization.
    \item A 1.4x reduction in MB emissions intensity (gCO$_2$e/kWh) of Google's data center electricity from workload location impacts and clean energy procurement.
    \item A 36x reduction in Scope 1+3 emissions per prompt driven by lower machine-hours per prompt, and the associated reduction in amortized embodied emissions.
\end{enumerate}
}

We expect these dramatic emissions-efficiency gains to come from a combination of efforts, including:
\begin{itemize}
   \item \textit{Smarter model architectures:} Gemini models are built on the Transformer model architecture~\citep{vaswani2017attention}, which provided a 10-100x efficiency boost over the previous state-of-the-art architectures for language modeling~\citep{kaplan2020scaling}. We design models with inherently efficient structures and techniques like Mixture-of-Experts (MoE) and hybrid reasoning. MoE, for example, allows us to activate a small subset of a large model specifically required to respond to a prompt, reducing computations and data transfer by a factor of 10-100x.  We also make use of more efficient implementations of the attention computation than those described in the original Transformer paper.
   \item \textit{Efficient algorithms \& quantization:} We continuously refine the algorithms that power our models with methods like \textit{Accurate Quantized Training} (AQT)~\citep{aqt2022github} that uses narrower data types to maximize efficiency and reduce energy consumption for serving without compromising response quality.
   \item \textit{Optimized inference and serving:} We constantly improve AI models delivery for responsiveness and efficiency. Technologies like Speculative Decoding serve more responses with fewer AI accelerators~\citep{leviathan2023fast}. We use distillation~\citep{hinton2015distilling} to create more efficient, serving-optimized models (such as Gemini Flash and Flash-Lite) using our larger, more capable models as teachers.  As an added benefit of using smaller, serving-optimized models, we can push the batch size higher for serving while still meeting the same latency goals, further pushing up efficiency and hardware utilization.
   \item \textit{Custom-built hardware:} We design our TPUs from the ground up to give higher performance per watt. Our AI models and TPUs are co-designed, ensuring our software takes full advantage of our hardware. Our latest generation, Ironwood, is 30x more energy-efficient than our first publicly-available TPU. 
   \item \textit{Optimized idling:} Our serving stack makes highly efficient use of CPUs and minimizes accelerators idling by dynamically moving models based on demand in near-real-time, rather than a “set it and forget” approach. 
   \item \textit{ML software stack:} Our XLA ML compiler~\citep{openxla_xla}, Pallas kernels~\citep{jaxpallas}, and Pathways~\citep{barham2022pathways} systems enable model computations expressed in higher-level systems like JAX to be run efficiently on our accelerators serving hardware.
   \item \textit{Ultra-efficient data centers:} Google data centers are among the industry’s most efficient, operating at a fleet-wide average PUE of 1.09---only 9\% over zero overhead---and delivering over six times more computing power per unit of electricity than five years ago~\citep{Google_Data_Center_Efficiency}. Google continues to advance our 120\% replenishment goal to drive net water consumption to zero, and optimize our cooling systems---balancing the local trade-off between energy, water, and emissions by conducting science-backed watershed health assessments to guide cooling type selection and limit water use in high-stress locations. 
   \item \textit{Clean energy procurement:} Google continues to procure clean energy generation in pursuit of our 24/7 carbon-free ambition. In the 2025 annual Environmental Report~\citep{google2025environmental}, we have shown that despite continued growth of electricity consumption from 2023 to 2024, Scope 2 MB emissions have decreased over the same timeframe. This demonstrates an important decoupling between electricity consumption and emissions impact for Google's data centers -- decreasing Google's fleetwide Scope 2 MB emissions factor by 30\% from 2023 to 2024.
\end{itemize}

\section{Conclusions}
The proliferation of large-scale AI products necessitates a transparent and comprehensive understanding of their AI serving environmental footprint. This paper addresses a critical gap in the field by proposing and applying a comprehensive, full-stack methodology for measuring the energy consumption, carbon emissions, and water consumption of AI inference in a live production environment.

Our primary finding is that the environmental impact of AI serving can be significantly underestimated by existing, narrower measurement approaches. For Google's Gemini Apps products, a median text prompt consumes 0.24 Wh of energy, generates 0.03 gCO$_2$e, and consumes 0.26 mL of water. These figures are more comprehensive than many previously published metrics, but also end up being one or two orders of magnitude smaller than many existing estimates or measurements of AI inference benchmarks. We interpret this difference to come from three factors:
\begin{enumerate}
    \item In-situ measurement of energy consumption will be a more precise representation of actual energy consumption, and this study uses primary data on user prompt volumes.
    \item Existing measurements of AI inference energy consider open-source models that are likely not at the Pareto frontier of performance efficiencies.
    \item Deployment of AI inference in a production environment may be more efficient than benchmark experiments due to efficient batching of prompts at scale.
\end{enumerate}

The implications of this work are twofold. First, it establishes that for environmental metrics to be actionable and comparable across different models and providers, it is critical to consider a standardized, comprehensive measurement boundary. Without it, reported figures can vary by orders of magnitude for similar tasks, hindering transparency and accountability. Second, this holistic approach provides the necessary visibility to hotspot and properly incentivize efficiency gains across the entire AI serving stack. Optimizing for a comprehensive metric can drive substantial environmental improvements, highlighted by the 33x reduction in the median Gemini Apps’ per-prompt energy use \textcolor{black}{and 44x reduction in per-prompt emissions over the past year}.

While the impact of a single prompt is low compared to many daily activities, the immense scale of user adoption globally means that continued focus on reducing the environmental cost of AI is imperative. We advocate for the widespread adoption of this or similarly comprehensive measurement frameworks to ensure that as the capabilities of AI advance, their environmental efficiency does as well. We hope that this study contributes to ongoing efforts to develop efficient AI at this critical time to address human-induced climate impacts, and nascent time for AI capabilities.

%%%%%%%%%%%% Supplementary Methods %%%%%%%%%%%%
%\footnotesize
%\section*{Methods}

%%%%%%%%%%%%% Acknowledgements %%%%%%%%%%%%%
\footnotesize
\section*{Acknowledgements}
Thank you to our colleagues at Google who helped shape and improve this research and the resulting paper. Thank you especially to David Culler, Rachana Fellinger, Chase Hensel, Tom Lewin, Shannon Lane, Jiyang Li, and Sashidhar Battu who helped shape the direction of the research and provided critical feedback. Thank you to David Culler, Vincent Poncet, Chrissy Patterson, Sushma Prasad, and Owen Lehmer for feedback on the draft.

%%%%%%%%%%%%%%   Bibliography   %%%%%%%%%%%%%%
\footnotesize
\bibliography{references}

%%%%%%%%%%%%%%   Appendix   %%%%%%%%%%%%%%
\footnotesize
\section*{Appendix}
\subsection*{Appendix A: Normalization}
The factors impacting overhead energy consumption, water consumption, and emissions typically vary over time, and by location, climatic condition, cooling technology, and other external variables. 
\begin{itemize}
    \item \textit{Overhead Energy Consumption:} Power demand from cooling systems varies significantly based on technology and local climate. Generally, air-cooled systems are less energy efficient than adiabatic or evaporative cooled systems. Regardless of technology, power consumption increases during hotter and more humid periods. 
    \item \textit{Water Consumption:} The data center’s water demand varies significantly with cooling technology and climatic conditions. Air-cooled systems consume little to no water, while adiabatic or evaporative systems consume increasingly more water during hotter and drier periods. 
    \item \textit{Emission Factors:} Scope 2 emissions vary based on seasonal patterns, reflecting changes in energy demand and the grid’s mix of power generation sources.  
\end{itemize}

This variability can skew benchmarking and trends. For example, an AI model running during the winter season in optimal cooling conditions may appear more efficient due to lower overhead energy and water consumption. Similarly, an AI model running in an advanced or optimally sited data center will have a smaller footprint than the same model running in a less modern or suboptimally sited data center. 

To ensure accurate AI model benchmarking and comparison, Power Usage Effectiveness (PUE), Water Usage Effectiveness (WUE), and grid emission factors should be applied on a trailing-twelve-month (TTM) or annual-average basis. To normalize for site-specific variations, the fleetwide average weighted by the energy serving AI compute should be used. We did both of these things. For example, this means that a May 2024 estimate uses May 2024-specific numbers for energy/query, while using annual fleetwide average numbers for PUE, WUE, and grid emissions factors. This ensures that the data accurately reflect the operational profile of the entire data center fleet serving AI during typical operation, and not a single facility in an optimal climate.

%%%%%%%%%%%%%%%%   End   %%%%%%%%%%%%%%%%
%\end{multicols}  % Method B for two-column formatting (doesn't play well with line numbers), comment out if using method A
\end{document}